\documentclass{article}

\usepackage{times}
\usepackage{graphicx} 
\usepackage{subcaption}

\PassOptionsToPackage{numbers,compress}{natbib}

\usepackage{bbm}

\usepackage{algorithm}
\usepackage{algorithmic}

\usepackage{multirow}
\usepackage{rotating}
\usepackage{booktabs}

\usepackage[olditem,oldenum]{paralist}

\usepackage{symbols}

\usepackage[pagebackref=true,breaklinks=true,bookmarks=false]{hyperref}

\usepackage[accepted]{whi2016}

\icmltitlerunning{Human Attention in VQA: Do Humans and Deep Networks Look at the Same Regions?}

\begin{document}

\twocolumn[
\icmltitle{Human Attention in Visual Question Answering:\\
           Do Humans and Deep Networks Look at the Same Regions?}

\icmlauthor{Abhishek Das\thanks{\,\, Denotes equal contribution.}\textsuperscript{$\dagger$}}{abhshkdz@vt.edu}

\icmlauthor{Harsh Agrawal\footnotemark[1]\textsuperscript{$\dagger$}}{harsh92@vt.edu}

\icmlauthor{C. Lawrence Zitnick\textsuperscript{$\ddagger$}}{zitnick@fb.com}

\icmlauthor{Devi Parikh\textsuperscript{$\dagger$}}{parikh@vt.edu}

\icmlauthor{Dhruv Batra\textsuperscript{$\dagger$}}{dbatra@vt.edu}
\icmladdress{\textsuperscript{$\dagger$}Virginia Tech, Blacksburg\\
\textsuperscript{$\ddagger$}Facebook AI Research, Menlo Park}

\icmlkeywords{visual question answering, attention, human}

\vskip 0.3in
]

\footnotetext{Denotes equal contribution.}

\begin{abstract}
  We conduct large-scale studies on `human attention' in Visual Question Answering (VQA) to understand where humans choose to look to answer questions about images.
We design and test multiple game-inspired novel attention-annotation interfaces that require the subject to sharpen regions of a blurred image to answer a question.
Thus, we introduce the VQA-HAT (Human ATtention) dataset.
We evaluate attention maps generated by state-of-the-art VQA models against human attention both qualitatively (via visualizations) and quantitatively (via rank-order correlation).
Our experiments show that current attention models in VQA do not seem to be looking at the same regions as humans.
\end{abstract}

\section{Introduction}
\label{sec:intro}

It helps to pay attention.
Humans have the ability to quickly perceive a scene by selectively attending to parts of the image instead of processing the whole scene in its entirety \cite{rensink_vc2000}.
Inspired by human attention, a recent trend in computer vision and deep learning is to build computational models of attention.
Given an input signal, these models learn to attend to parts of it for further processing and have been successfully applied in machine translation \cite{bahdanau_iclr15,firat_arxiv16}, object recognition \cite{ba_iclr15,mnih_arxiv14,sermanet_arxiv14}, image captioning \cite{xu_arxiv15,cho_arxiv15} and visual question answering \cite{yang_arxiv15,lu_arxiv16,hxu_arxiv15,xiong_arxiv16}.

\begin{figure}[t]
    \includegraphics[width=1\linewidth]{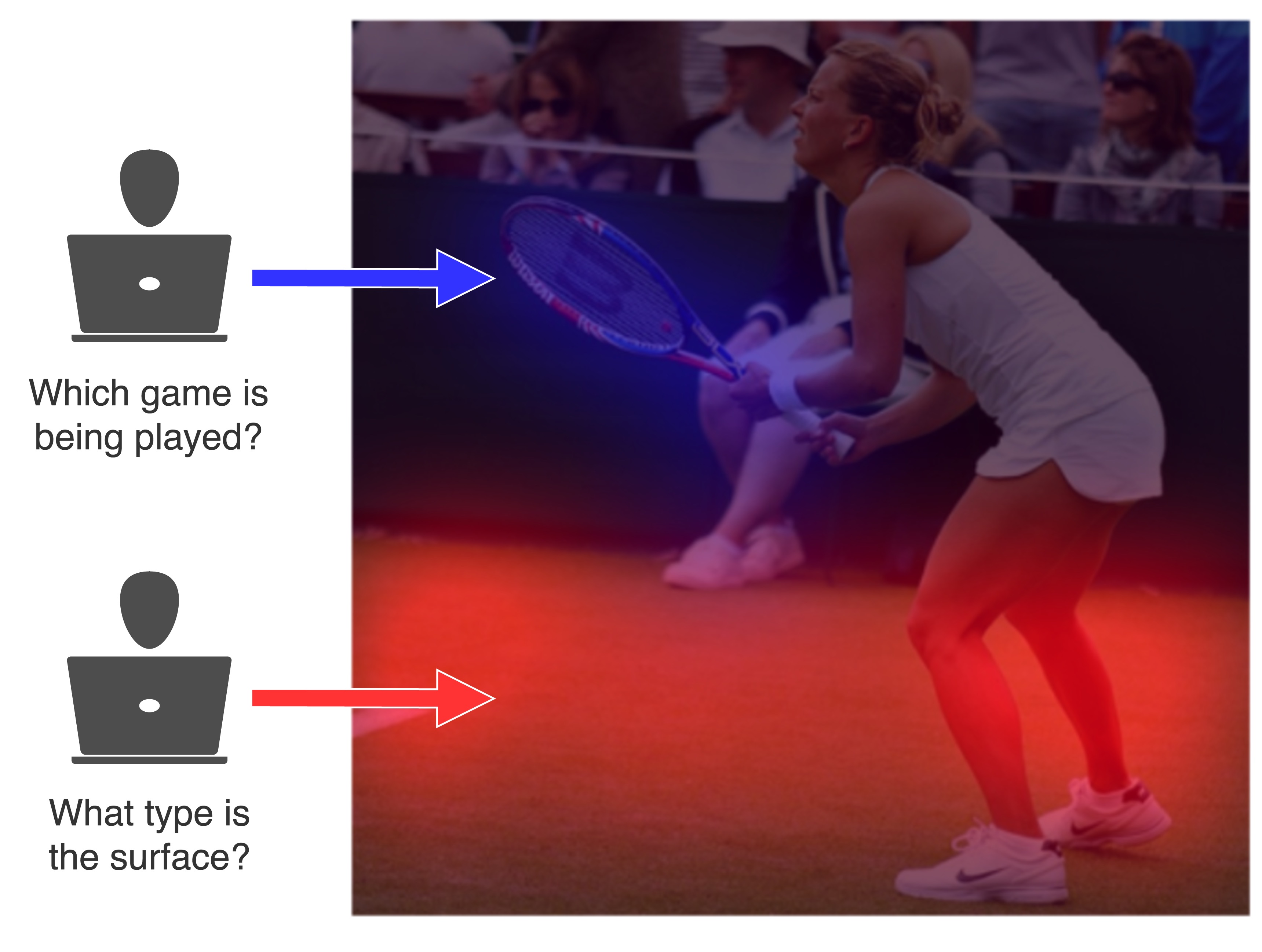}
    \caption{Different human attention regions based on question (best viewed in color).}
    \label{fig:teaser}
\end{figure}

In this work, we study attention for the task of Visual Question Answering (VQA).
Unlike image captioning, where a coarse understanding of an image is often sufficient for producing generic descriptions \cite{devlin_arxiv15}, visual questions selectively target different areas of an image including background details and underlying context.
This suggests that a VQA model may benefit from an explicit or implicit attention mechanism to answer a question correctly.

\begin{figure*}[ht]
  \centering
  \begin{subfigure}[b]{0.32\textwidth}
    \includegraphics[width=1\linewidth]{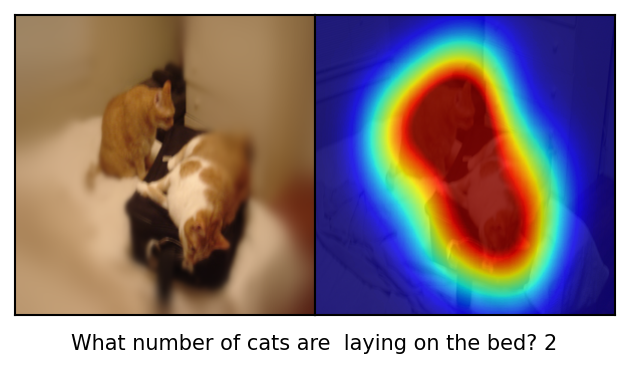}
    \caption{}
  \end{subfigure}
  \begin{subfigure}[b]{0.32\textwidth}
    \includegraphics[width=1\linewidth]{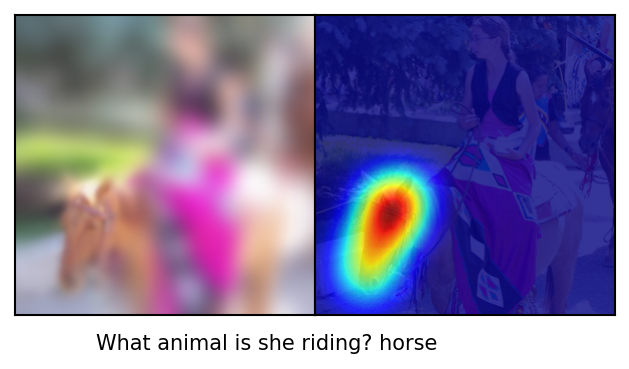}
    \caption{}
  \end{subfigure}
  \begin{subfigure}[b]{0.32\textwidth}
    \includegraphics[width=1\linewidth]{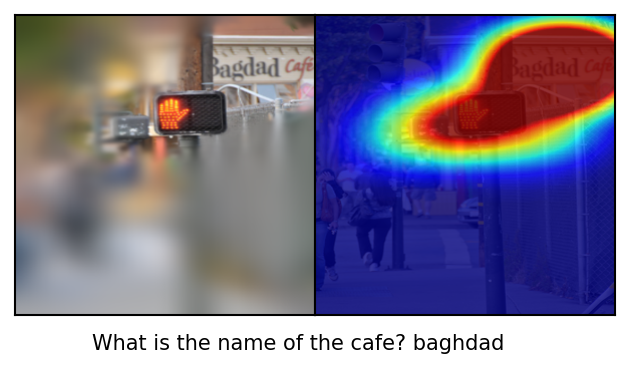}
    \caption{}
  \end{subfigure}
  \caption{
    (a-c): Column 1 shows deblurred image, and column 2 shows human attention map.}
  \label{fig:human_maps}
\end{figure*}


In this work, we are interested in the following questions:
1) Which image regions do humans choose to look at in order to answer questions about images?
2) Do deep VQA models with attention mechanisms attend to the same regions as humans?

We design and conduct studies to collect ``human attention maps".
\figref{fig:teaser} shows human attention maps on the same image for two different questions.
When asked `What type is the surface?', humans choose to look at the floor, while attention for `Which game is being played?' is concentrated around the player and racket.


These human attention maps can be used both for evaluating machine-generated attention maps and for explicitly training attention-based models.

\noindent \textbf{Contributions.}
First, we design and test multiple game-inspired novel interfaces for collecting human attention maps of where humans choose to look to answer questions from the large-scale VQA dataset \cite{antol_iccv15}; this VQA-HAT (Human ATtention) dataset will be released publicly.
Second, we perform qualitative and quantitative comparison of the maps generated by state-of-the-art attention-based VQA models \cite{yang_arxiv15,lu_arxiv16} and a task-independent saliency baseline \cite{judd_iccv09} against our human attention maps through visualizations and rank-order correlation.
We find that machine-generated attention maps from the most accurate VQA model have a mean rank-correlation of 0.26 with human attention maps, which is worse than task-independent saliency maps that have a mean rank-correlation of 0.49.
It is well understood that task-independent saliency maps have a `center bias'~\cite{judd_iccv09}.
After we control for this center bias in our human attention maps, we find that the correlation of task-independent saliency is poor (as expected), while trends for machine-generated VQA-attention maps remain the same (which is promising).

\section{Related Work}
\label{sec:related}





\begin{figure*}[h!t]
    \centering
    \begin{subfigure}[b]{0.32\textwidth}
        \includegraphics[width=1\textwidth]{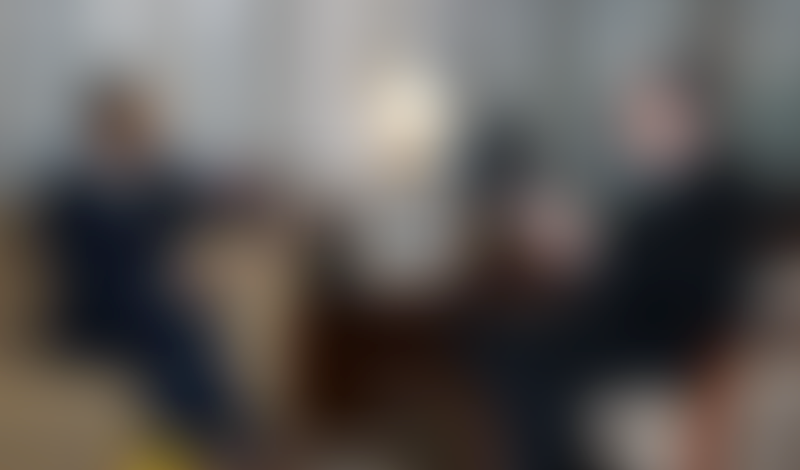}%
        \caption{Initial blurred image}
        \label{fig:step1}
    \end{subfigure}
    \begin{subfigure}[b]{0.32\textwidth}%
        \includegraphics[width=1\textwidth]{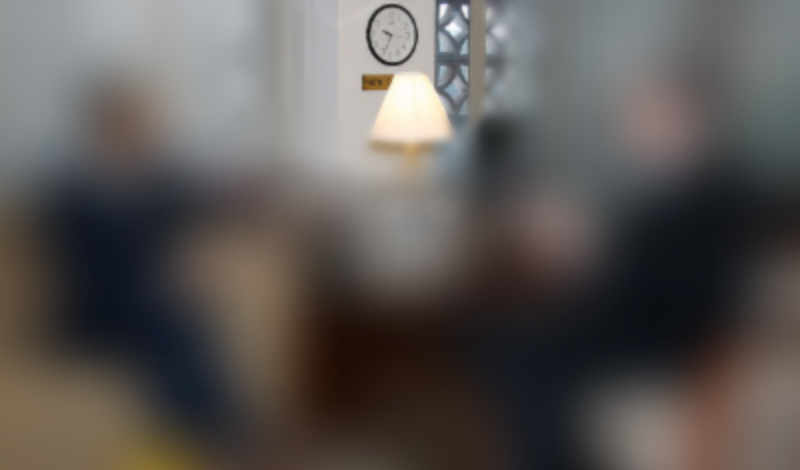}
        \caption{Regions sharpened by subject}
        \label{fig:step2}
    \end{subfigure}
    \begin{subfigure}[b]{0.32\textwidth}
        \includegraphics[width=1\textwidth]{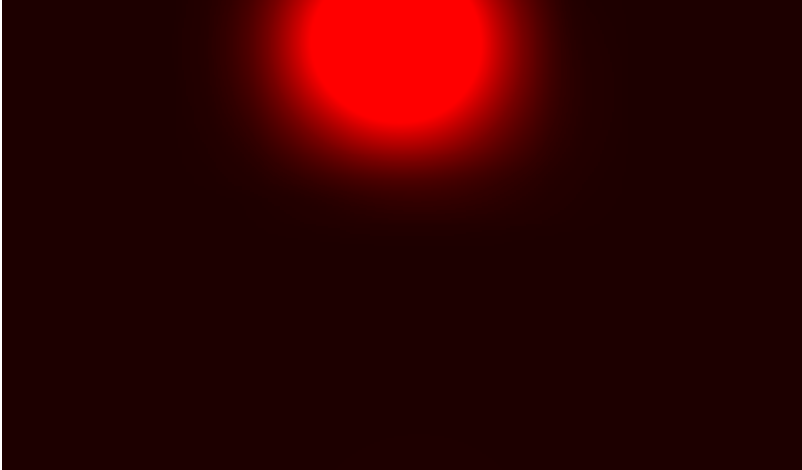}%
        \caption{Attention map}
        \label{fig:step3}
    \end{subfigure}
    \caption{Deblurring procedure to collect attention maps.
    } \vspace{-10pt}
\label{fig:task_steps}
\end{figure*}

Our work draws on recent work in attention-based VQA and human studies in saliency prediction.
We work with the free-form and open-ended VQA dataset released by \cite{antol_iccv15}.


\noindent \textbf{VQA Models.} Attention-based models for VQA typically use convolutional neural networks to highlight relevant regions of image given a question.
Stacked Attention Networks (SAN) proposed in \cite{yang_arxiv15} use LSTM encodings of question words to produce a spatial attention distribution over the convolutional layer features of the image.
Hierarchical Co-Attention Network \cite{lu_arxiv16} generates multiple levels of image attention based on words, phrases and complete questions, and is the top entry on the VQA Challenge\footnote{\url{http://visualqa.org/challenge.html}} as of the time of this submission.
Another interesting approach uses question parsing to compose the neural network from modules, attention being one of the sub-tasks addressed by these modules \cite{andreas_arxiv16}.

Note that all these works are \emph{unsupervised} attention models, where ``attention'' is simply an intermediate variable (a spatial distribution) that is produced by the model to optimize downstream loss (VQA cross-entropy).
The fact that some (it's unclear how many) of these spatial distributions end up being interpretable is simply fortuitous.
In contrast, we study where humans choose to look to answer visual questions.
These human attention maps can be used to evaluate unsupervised maps.

\noindent \textbf{Human Studies.} There's a rich history of work in collecting eye tracking data from human subjects to gain an understanding of image saliency and visual perception \cite{jiang_eccv14,judd_iccv09,feifei_jov07,yarbus_1967}.
Eye tracking data to study natural visual exploration \cite{jiang_eccv14,judd_iccv09} is useful but difficult and expensive to collect on a large scale.
\cite{jiang_cvpr15} established mouse tracking as an accurate approach to collecting attention maps.
They collected large-scale attention annotations for MS COCO \cite{coco} on Amazon Mechanical Turk (AMT).
While \cite{jiang_cvpr15} studies natural exploration and collects task-independent human annotations by asking subjects to freely move the mouse cursor to anywhere they wanted to look on a blurred image, our approach is task-driven.


Specifically, as described in section \ref{sec:dataset}, we collect ground truth attention annotations by instructing subjects to sharpen parts of a blurred image that are important for answering the questions accurately.
Section \ref{sec:experiments} covers evaluation of unsupervised attention maps generated by VQA models against our human attention maps.

\section{VQA-HAT (Human ATtention) Dataset}
\label{sec:dataset}
We design and test multiple game-inspired novel interfaces for conducting large-scale human studies on AMT.
Our basic interface design consists of a ``deblurring" exercise for answering visual questions.
Specifically, we present subjects with a blurred image and a question about the image, and ask subjects to sharpen regions of the image that will help them answer the question correctly, in a smooth, click-and-drag, `coloring' motion with the mouse.
Successively scrubbing the same region progressively sharpens it.
\figref{fig:task_steps} shows intermediate steps in our attention annotation interface, from a completely blurry image to a deblurred attention map.

\textbf{Dataset Evaluation.} We ran pilot studies on AMT to experiment with multiple interfaces -- where we ask just the question, or show both the question and answer, or question, answer as well as original high-resolution image, along with blurred image.
In order to quantitatively evaluate the interfaces, we conducted a second human study where (a second set of) subjects where shown the attention-sharpened images generated from each of the attention interfaces from the first experiment and asked to answer the question.
The intuition behind this experiment is that if the attention map revealed too little information, this second set of subjects would answer the question incorrectly.
Table~\ref{tab:human_acc} shows VQA accuracies of the answers given by human subjects under these 3 interfaces.
We can see that the ``Blurred Image with Answer" interface gives the highest accuracy on evaluation by humans.

Since the payments structure on AMT encourage completing tasks as quickly as possible, this implicitly incentivizes subjects to deblur as few regions as possible, and our human study shows that humans can still answer questions.
Thus, overall we achieve a balance between highlighting too little or too much.


\begin{table}[h]
\centering
\resizebox{0.98\columnwidth}{!}{%
\begin{tabular}{ccc}
\toprule
Interface Type                          & Human Accuracy \\ \midrule
Blurred Image without Answer            & 75.2        \\
Blurred Image with Answer               & 78.7        \\
Blurred \& Original Image with Answer   & 71.2         \\
Original Image                          & 80.0         \\ \bottomrule
\end{tabular}%
}
\caption{Human accuracies to compare the quality of human attention maps collected by different interfaces.
}
\label{tab:human_acc}
\end{table}

We collected human attention maps for 58475 train (out of 248349 total) and 1374 val (out of 121512 total) question-image pairs in the VQA dataset.
Overall, we conducted approximately 20000 Human Intelligence Tasks (HITs) on AMT, among 800 unique workers.
\figref{fig:human_maps} shows examples of collected human attention maps.
This VQA-HAT dataset will be released publicly.

\begin{figure}[h!tp]
    \includegraphics[width=1\linewidth]{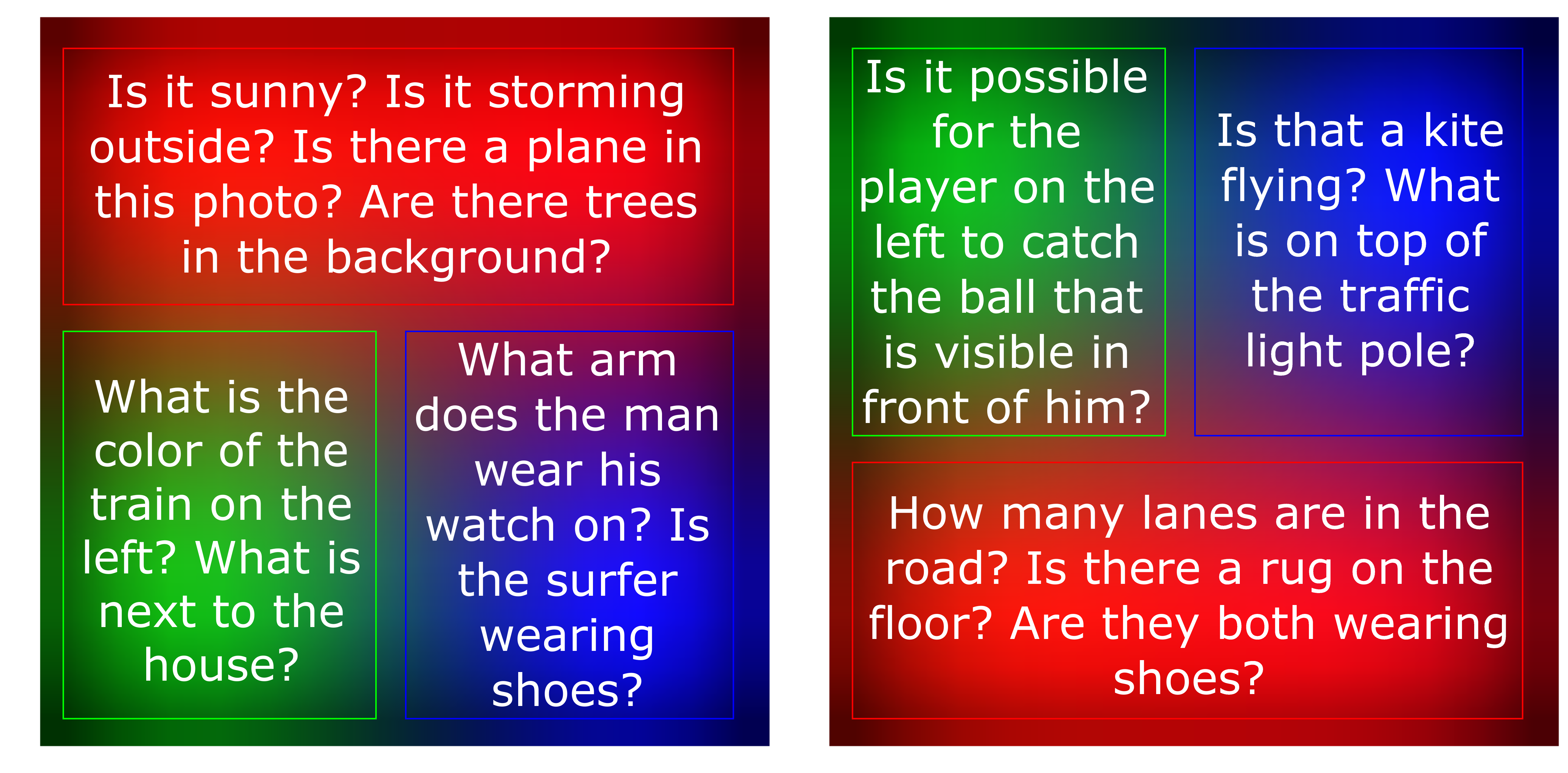}
    \caption{
    }
    \label{fig:analysis}
\end{figure}

To visualize the collected dataset, we cluster the human attention maps and visualize the average attention map and example questions falling in each of them for 6 selected clusters in \figref{fig:analysis}.

\section{Human Attention Maps vs Unsupervised Attention Models}
\label{sec:experiments}


Now that we have collected human attention maps, we can ask the question -- do unsupervised attention models learn to predict attention maps that are similar to human attention maps?
To rephrase, \emph{do neural networks look at the same regions as humans to answer a visual question?}

\noindent \textbf{VQA Attention Models.} We evaluate maps generated by the following unsupervised models:
\begin{compactitem}
    \item Stacked Attention Network (SAN) \cite{yang_arxiv15} with two attention layers (SAN-2)\footnote{Code available at \url{https://github.com/zcyang/imageqa-san}.}.
    \item Hierarchical Co-Attention Network (HieCoAtt) \cite{lu_arxiv16} with word-level (HieCoAtt-W), phrase-level (HieCoAtt-P) and question-level (HieCoAtt-Q) attention maps; we evaluate all three maps\footnote{Code available at \url{https://github.com/jiasenlu/HieCoAttenVQA}}.
\end{compactitem}


\noindent \textbf{Comparison Metric: Rank Correlation.} We first scale both the machine-generated and human attention maps to 14x14, rank the pixels according to their spatial attention and then compute correlation between these two ranked lists.
We choose an order-based metric so as to make the evaluation invariant to absolute spatial probability values which can be made peaky or diffuse by tweaking a `temperature' parameter.

\begin{table}[h]
\resizebox{0.98\columnwidth}{!}{%
\begin{tabular}{ccc}
\toprule
Model & Rank-correlation \\ \midrule
SAN-2 \cite{yang_arxiv15}  & 0.249 $\pm$ 0.004 \\ \cmidrule(lr){1-2}
HieCoAtt-W \cite{lu_arxiv16} & 0.246 $\pm$ 0.004 \\
HieCoAtt-P \cite{lu_arxiv16} & 0.256 $\pm$ 0.004 \\
HieCoAtt-Q \cite{lu_arxiv16} & 0.264 $\pm$ 0.004 \\ \cmidrule(lr){1-2}
Random & 0.000 $\pm$ 0.001 \\ \cmidrule(lr){1-2}
Judd et al. \cite{judd_iccv09} & 0.497 $\pm$ 0.004 \\ \cmidrule(lr){1-2}
Human & 0.623 $\pm$ 0.003 \\ \bottomrule
\end{tabular}
}
\caption{Mean rank-correlation coefficients (higher is better); error bars show standard error of means.}
\label{tab:eval}
\end{table}

\noindent Table~\ref{tab:eval} shows rank-order correlation averaged over all image-question pairs on the validation set.
We compare with random attention maps and task-independent saliency maps generated by a model trained to predict human eye fixation locations where subjects are asked to freely view an image for 3 seconds \cite{judd_iccv09}.
Both SAN-2 and HieCoAtt attention maps are positively correlated with human attention maps, but not as strongly as task-independent Judd saliency maps.
Our findings lead to two take-away messages with significant potential impact on future research in this active field.
First, current VQA attention models do not seem to be `looking' at the same regions as humans to produce an answer.
Second, as attention-based VQA models become more accurate ($58.9\%$ SAN $\rightarrow$ $62.1\%$ HieCoAtt), they seem to be (slightly) better correlated with humans in terms of where they look.
Our dataset will allow for a more thorough validation of this observation as future attention-based VQA models are proposed.
To put these numbers in perspective, we computed inter-human agreement on the validation set by collecting 3 human attention maps per image-question pair and computing mean rank-correlation, which is 0.623.



\noindent \textbf{Center Bias.} Judd saliency maps aim to predict human eye fixations during natural visual exploration.
These tend to have a strong center bias \cite{judd_iccv09}.
Although our human attention maps dataset is not an eye tracking study, the center bias still exists albeit not as severe.
One potential source of this center bias is the fact that the VQA dataset was human-generated by subjects looking at the images.
Thus, salient objects in the center of the image are likely be potential subjects of the questions.
We compute rank-correlation of a synthetically generated central attention map with Judd saliency and human attention maps.
Judd saliency maps have a mean rank-correlation of 0.877 and human attention maps have a mean rank-correlation of 0.458 on the validation set.


\begin{table}[h!]
\resizebox{0.98\columnwidth}{!}{%
\begin{tabular}{ccc}
\toprule
Model & Rank-correlation \\ \midrule
SAN-2 \cite{yang_arxiv15}  & 0.038 $\pm$ 0.011 \\ \cmidrule(lr){1-2}
HieCoAtt-W \cite{lu_arxiv16} & 0.062 $\pm$ 0.012 \\
HieCoAtt-P \cite{lu_arxiv16} & 0.048 $\pm$ 0.010 \\
HieCoAtt-Q \cite{lu_arxiv16} & 0.114 $\pm$ 0.012 \\ \cmidrule(lr){1-2}
Judd et al. \cite{judd_iccv09} & -0.063 $\pm$ 0.009 \\ \bottomrule
\end{tabular}
}
\caption{Mean rank-correlation coefficients (higher is better) on the reduced set without center bias; error bars show standard error of means.
}
\label{tab:eval_centerless}
\end{table}

To eliminate the effect of center bias in this evaluation, we removed human attention maps that have a positive rank-correlation with the center attention map.
We compute rank-correlation of machine-generated attention with human attention on this reduced set.
See Table~\ref{tab:eval_centerless}.
Mean correlation goes down significantly for Judd saliency maps since they have a strong center bias.
Relative trends among SAN-2 \& HieCoAtt are similar to those over the whole validation set (reported in Table~\ref{tab:eval}).
HieCoAtt-Q now has a higher correlation with human attention maps than Judd saliency.
This demonstrates that discounting the center bias, VQA-specific machine attention maps correlate better with VQA-specific human attention maps than task independent machine saliency maps.


\section{Conclusion \& Discussion}
\label{sec:discussion}



We introduce and release the VQA-HAT dataset.
This dataset can be used to evaluate attention maps generated in an unsupervised manner by attention-based VQA models, or to explicitly train models with attention supervision for VQA.
We quantify whether current attention-based VQA models are `looking' at the same regions of the image as humans do to produce an answer.

\noindent \textbf{Necessary vs Sufficient Maps.} Are human attention maps `necessary' and/or `sufficient'?
If regions highlighted by the human attention maps are sufficient to answer the question accurately, then so is any region that is a superset.
For example, if attention mass is concentrated on a `cat' for `What animal is present in the picture?', then an attention map that assigns weights to any arbitrary-sized region that includes the `cat' is sufficient as well.
On the contrary, a \emph{necessary} and sufficient attention map would be the smallest visual region sufficient for answering the question accurately.
It is an ill-posed problem to define a necessary attention map in the space of pixels; random pixels can be blacked out and chances are that humans would still be able to answer the question given the resulting subset attention map.
Our work thus poses an interesting question for future work -- what is the right \emph{semantic} space in which it is meaningful to talk about necessary and sufficient attention maps for humans?

\textbf{Acknowledgements.} We thank Jiasen Lu and Ramakrishna Vedantam for helpful suggestions and discussions.
This work was supported in part by the following:
National Science Foundation CAREER awards to DB and DP,
Army Research Office YIP awards to DB and DP,
ICTAS Junior Faculty awards to DB and DP,
Army Research Lab grant W911NF-15-2-0080 to DP and DB,
Office of Naval Research grant N00014-14-1-0679 to DB,
Paul G. Allen Family Foundation Allen Distinguished Investigator award to DP,
Google Faculty Research award to DP and DB,
AWS in Education Research grant to DB, and NVIDIA GPU donation to DB.

{\small
\bibliographystyle{icml2016}
\bibliography{main}

\begin{thebibliography}{21}
\providecommand{\natexlab}[1]{#1}
\providecommand{\url}[1]{\texttt{#1}}
\expandafter\ifx\csname urlstyle\endcsname\relax
  \providecommand{\doi}[1]{doi: #1}\else
  \providecommand{\doi}{doi: \begingroup \urlstyle{rm}\Url}\fi

\bibitem[Andreas et~al.(2016)Andreas, Rohrbach, Darrell, and
  Klein]{andreas_arxiv16}
Andreas, Jacob, Rohrbach, Marcus, Darrell, Trevor, and Klein, Dan.
\newblock Learning to compose neural networks for question answering.
\newblock \emph{CoRR}, abs/1601.01705, 2016.
\newblock URL \url{http://arxiv.org/abs/1601.01705}.

\bibitem[Antol et~al.(2015)Antol, Agrawal, Lu, Mitchell, Batra, Zitnick, and
  Parikh]{antol_iccv15}
Antol, Stanislaw, Agrawal, Aishwarya, Lu, Jiasen, Mitchell, Margaret, Batra,
  Dhruv, Zitnick, C.~Lawrence, and Parikh, Devi.
\newblock Vqa: Visual question answering.
\newblock In \emph{ICCV}, 2015.

\bibitem[Ba et~al.(2015)Ba, Mnih, and Kavukcuoglu]{ba_iclr15}
Ba, Jimmy~Lei, Mnih, Volodymyr, and Kavukcuoglu, Koray.
\newblock {Multiple Object Recognition With Visual Attention}.
\newblock \emph{Iclr-2015}, 2015.

\bibitem[Bahdanau et~al.(2014)Bahdanau, Cho, and Bengio]{bahdanau_iclr15}
Bahdanau, Dzmitry, Cho, Kyunghyun, and Bengio, Yoshua.
\newblock Neural machine translation by jointly learning to align and
  translate.
\newblock \emph{CoRR}, abs/1409.0473, 2014.
\newblock URL \url{http://arxiv.org/abs/1409.0473}.

\bibitem[Cho et~al.(2015)Cho, Courville, and Bengio]{cho_arxiv15}
Cho, KyungHyun, Courville, Aaron~C., and Bengio, Yoshua.
\newblock Describing multimedia content using attention-based encoder-decoder
  networks.
\newblock \emph{CoRR}, abs/1507.01053, 2015.
\newblock URL \url{http://arxiv.org/abs/1507.01053}.

\bibitem[Devlin et~al.(2015)Devlin, Gupta, Girshick, Mitchell, and
  Zitnick]{devlin_arxiv15}
Devlin, Jacob, Gupta, Saurabh, Girshick, Ross, Mitchell, Margaret, and Zitnick,
  C.~Lawrence.
\newblock {Exploring Nearest Neighbor Approaches for Image Captioning}.
\newblock \emph{arXiv preprint}, 2015.

\bibitem[Fei-Fei et~al.(2007)Fei-Fei, Iyer, Koch, and Perona]{feifei_jov07}
Fei-Fei, Li, Iyer, Asha, Koch, Christof, and Perona, Pietro.
\newblock What do we perceive in a glance of a real-world scene?
\newblock \emph{Journal of Vision}, 7\penalty0 (1):\penalty0 10, 2007.
\newblock \doi{10.1167/7.1.10}.
\newblock URL \url{+ http://dx.doi.org/10.1167/7.1.10}.

\bibitem[Firat et~al.(2016)Firat, Cho, and Bengio]{firat_arxiv16}
Firat, Orhan, Cho, KyungHyun, and Bengio, Yoshua.
\newblock Multi-way, multilingual neural machine translation with a shared
  attention mechanism.
\newblock \emph{CoRR}, abs/1601.01073, 2016.
\newblock URL \url{http://arxiv.org/abs/1601.01073}.

\bibitem[Jiang et~al.(2014)Jiang, Xu, and Zhao]{jiang_eccv14}
Jiang, Ming, Xu, Juan, and Zhao, Qi.
\newblock {Saliency in Crowd}.
\newblock \emph{ECCV}, 2014.

\bibitem[Jiang et~al.(2015)Jiang, Huang, Duan, and Zhao]{jiang_cvpr15}
Jiang, Ming, Huang, Shengsheng, Duan, Juanyong, and Zhao, Qi.
\newblock Salicon: Saliency in context.
\newblock In \emph{CVPR}, June 2015.

\bibitem[Judd et~al.(2009)Judd, Ehinger, Durand, and Torralba]{judd_iccv09}
Judd, Tilke, Ehinger, Krista, Durand, Fr{\'e}do, and Torralba, Antonio.
\newblock Learning to predict where humans look.
\newblock In \emph{ICCV}, 2009.

\bibitem[Lin et~al.(2014)Lin, Maire, Belongie, Hays, Perona, Ramanan, Dollár,
  and Zitnick]{coco}
Lin, Tsung-Yi, Maire, Michael, Belongie, Serge, Hays, James, Perona, Pietro,
  Ramanan, Deva, Dollár, Piotr, and Zitnick, C.~Lawrence.
\newblock Microsoft {COCO}: Common objects in context, 2014.

\bibitem[{Lu} et~al.(2016){Lu}, {Yang}, {Batra}, and {Parikh}]{lu_arxiv16}
{Lu}, J., {Yang}, J., {Batra}, D., and {Parikh}, D.
\newblock {Hierarchical Co-Attention for Visual Question Answering}.
\newblock \emph{ArXiv e-prints}, May 2016.

\bibitem[Mnih et~al.(2014)Mnih, Heess, Graves, and Kavukcuoglu]{mnih_arxiv14}
Mnih, Volodymyr, Heess, Nicolas, Graves, Alex, and Kavukcuoglu, Koray.
\newblock {Recurrent Models of Visual Attention}.
\newblock \emph{arXiv preprint}, 2014.

\bibitem[Rensink(2000)]{rensink_vc2000}
Rensink, Ronald~A.
\newblock The dynamic representation of scenes.
\newblock \emph{Visual Cognition}, 7\penalty0 (1-3):\penalty0 17--42, 2000.
\newblock \doi{10.1080/135062800394667}.
\newblock URL \url{http://dx.doi.org/10.1080/135062800394667}.

\bibitem[Sermanet et~al.(2014)Sermanet, Frome, and Real]{sermanet_arxiv14}
Sermanet, Pierre, Frome, Andrea, and Real, Esteban.
\newblock Attention for fine-grained categorization.
\newblock \emph{CoRR}, abs/1412.7054, 2014.
\newblock URL \url{http://arxiv.org/abs/1412.7054}.

\bibitem[Xiong et~al.(2016)Xiong, Merity, and Socher]{xiong_arxiv16}
Xiong, Caiming, Merity, Stephen, and Socher, Richard.
\newblock Dynamic memory networks for visual and textual question answering.
\newblock \emph{CoRR}, abs/1603.01417, 2016.
\newblock URL \url{http://arxiv.org/abs/1603.01417}.

\bibitem[Xu \& Saenko(2015)Xu and Saenko]{hxu_arxiv15}
Xu, Huijuan and Saenko, Kate.
\newblock Ask, attend and answer: Exploring question-guided spatial attention
  for visual question answering.
\newblock \emph{CoRR}, abs/1511.05234, 2015.
\newblock URL \url{http://arxiv.org/abs/1511.05234}.

\bibitem[Xu et~al.(2015)Xu, Ba, Kiros, Cho, Courville, Salakhutdinov, Zemel,
  and Bengio]{xu_arxiv15}
Xu, Kelvin, Ba, Jimmy, Kiros, Ryan, Cho, Kyunghyun, Courville, Aaron~C.,
  Salakhutdinov, Ruslan, Zemel, Richard~S., and Bengio, Yoshua.
\newblock Show, attend and tell: Neural image caption generation with visual
  attention.
\newblock \emph{CoRR}, abs/1502.03044, 2015.
\newblock URL \url{http://arxiv.org/abs/1502.03044}.

\bibitem[Yang et~al.(2015)Yang, He, Gao, Deng, and Smola]{yang_arxiv15}
Yang, Zichao, He, Xiaodong, Gao, Jianfeng, Deng, Li, and Smola, Alexander~J.
\newblock Stacked attention networks for image question answering.
\newblock \emph{CoRR}, abs/1511.02274, 2015.
\newblock URL \url{http://arxiv.org/abs/1511.02274}.

\bibitem[Yarbus(1967)]{yarbus_1967}
Yarbus, A.~L.
\newblock \emph{Eye Movements and Vision}.
\newblock Plenum. New York., 1967.

\end{thebibliography}
}

\end{document}